\documentclass[letterpaper, 10 pt, conference]{ieeeconf}  %

\IEEEoverridecommandlockouts                              %

\usepackage{amsmath}
\usepackage{amssymb}  %

\usepackage{bm}  %
\usepackage{graphicx}  %
\usepackage{subcaption} %
\usepackage{siunitx}
\usepackage{algpseudocode} %
\usepackage{algorithm} %
\usepackage{flushend} %
\usepackage[T1]{fontenc} %
\usepackage{wrapfig} %
\usepackage{tabularx}  %
\usepackage{tabto}  %

\usepackage[colorinlistoftodos,prependcaption,textsize=tiny]{todonotes}

\presetkeys%
    {todonotes}%
    {inline,backgroundcolor=green}{}
\usepackage[
        backend=biber,
        style=numeric,
        sorting=none,
        doi=false,
        isbn=false,
        url=false]{biblatex}

\newcommand{\fig}[1]{Fig.~\ref{#1}}
\newcommand{\sect}[1]{Section~\ref{#1}}

\newcommand{\tabl}[1]{Table~\ref{#1}}
\newcommand{\alg}[1]{Algorithm~\ref{#1}}

\title{\LARGE \bf
A Walking Space Robot for On-Orbit Satellite Servicing: The ReCoBot
}

\author{Stefan Scherzinger$^{\ddagger 1}$, Jakob Weinland$^{\ddagger 1}$, Robert Wilbrandt$^{1}$, Pascal Becker$^{1}$,\\ Arne Roennau$^{1}$ and R\"udiger Dillmann$^{1}$%
\thanks{
This work was funded by the Deutsches Zentrum für Luft- und Raumfahrt (DLR)
under grant number 50RA2010 (ReCoBot - ReConfiguring and Manipulation RoBot).  
\medskip
\newline
 ${}^{\ddagger}$ Equal contribution.
\newline
 ${}^{1}$ All authors are with FZI Research Center for Information Technology, Haid-und-Neu-Str. 10-14, 76131 Karlsruhe, Germany
        {\tt\small \{scherzinger, weinland, wilbrandt, pbecker, roennau, dillmann\}@fzi.de}}%
}

\begin{document}

\maketitle
\thispagestyle{empty}
\pagestyle{empty}

\begin{abstract}
A key factor in the economic efficiency of satellites is their availability in orbit.
Replacing standardized building blocks, such as empty fuel tanks or outdated electronic modules, could greatly extend
the satellites' lifetime.
This, however, requires flexible robots that can locomote on the surface of these satellites for optimal accessibility and manipulation.
This paper introduces ReCoBot, a 7-axis walking space manipulator for locomotion and manipulation.
The robot can connect to compatible structures with its symmetric ends
and provides interfaces for manual teleoperation and motion planning with a constantly changing base and tip.
We build on open-source robotics software and easily available components to
evaluate the overall concept with an early stage demonstrator.
The proposed manipulator has a length of 1.20~m and a weight of 10.4~kg
and successfully locomotes over a satellite mockup in our lab environment.

\end{abstract}

\section{Introduction}
\begin{figure}
        \centering
        \includegraphics[width=0.49\textwidth]{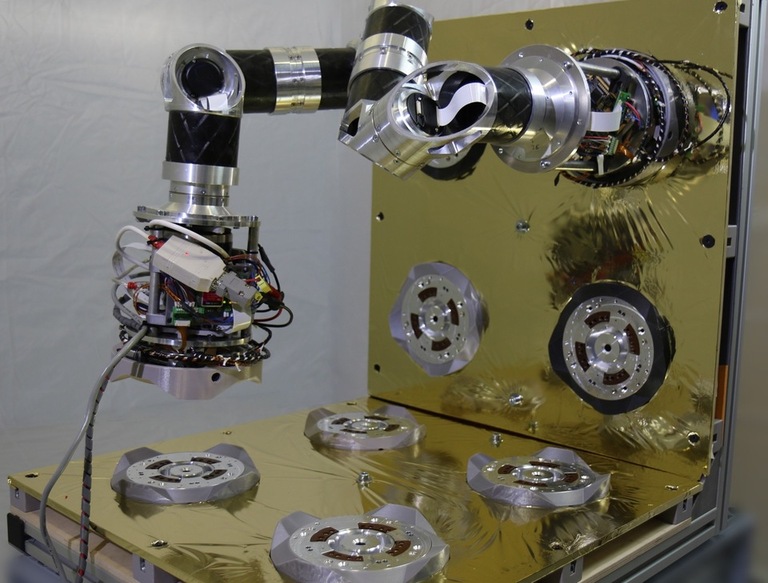}
        \caption{Our space manipulator ReCoBot, docked to a dummy satellite in our lab environment.
        The interfaces (grey) serve as waypoints for locomotion.
        }
        \label{fig:recobot_hanging}
\end{figure}

Thousands of satellites orbit the earth for communication and observation
and build the basis for key technologies of our modern life.
Despite functioning electronics, however, satellite lifetime often ends when running out of fuel.
And unforeseen events, such as damage through space debris, pose
an additional threat to their reliability.
The on-orbit servicing of satellites could partly compensate for those failures~\cite{Ellery2008},\cite{dubanchet2020eross},
but requires new concepts for satellite design and robotic servicers.
Fully modular approaches with box-shaped components~\cite{Kortman2015},
could simplify and reduce the on-orbit maintenance of spacecraft to the exchange of individual modules.
Building on a common interface for mechanical coupling and data exchange \cite{kreisel2019},
the challenge is to develop robotic manipulators that can dynamically reposition themselves on these spacecraft for on-orbit maintenance.
This paper presents ReCoBot,
a seven-axis, walking manipulator for the reconfiguration of satellite components.
Both of the robot's ends are equipped with the iSSI-interface~\cite{kreisel2019},
such that locomotion on compatible satellites is possible via dynamic changes of base and tip of the kinematic chain.
Following some concepts of \textit{new space}~\cite{Koechel2018},
we build ReCoBot with consumer electronics, light-weight robotic motors, and community-driven, open-source software.
This gives us the advantage of developing and realizing the concept in a short time
and obtaining early insights into the overall concept.
We address the challenges of designing such a robot both for space and for earth's gravity.
Although not required during its envisioned operation on-orbit,
the components need to sustain increased loads in earth's gravity for testing.
Furthermore, we build ReCoBot's software on the Robot Operating System (ROS)~\cite{Quigley2009},
which provides a rich set of tools and algorithms for robotics,
and we describe the adaptation needed for motion planning to realize our use case of caterpillar-like movement.
\fig{fig:recobot_hanging} shows ReCoBot in action.
Our proposed system
has a length of \SI{1.20}{m}, weighs \SI{10.4}{kg} and can hold
itself on vertical iSSI interfaces for locomotion.
With few further modifications, the design will be ready for a zero-gravity environment.

The structure of this paper is as follows:
\sect{sec:related_work} discusses related works with similar ideas and differences to our system,
and \sect{sec:recobot_overview} briefly summarizes our high-level goals for the next sections.
\sect{sec:hardware_design} then presents the hardware design, followed by electronics in
\sect{sec:electronics} and software in \sect{sec:software}.
Evaluation results are presented in \sect{sec:evaluation} before we
conclude the paper with perspectives on future work in
\sect{sec:conclusions}.

\section{Related Work}
\label{sec:related_work}
In the concept of on-orbit servicing,
the manipulators are usually co-developed with
a common standard for the connection of satellite modules.
In addition to the \textit{iSSI} interface on which we build in this paper,
there are further alternatives currently developed with similar purposes and characteristics.
Aligned with the \textit{iSSI}, they
comprise several functions in one multi-purpose interface
to meet the needs of future space applications, such as
mechanical coupling of components, power, fluid transfer, or data exchange~\cite{vinals2020standard},\cite{letier2020hotdock}.
These interfaces are then envisioned to connect
manipulators for operating on the surfaces of satellites.
Recent manipulators include the MOSAR-WM~\cite{deremetz2020mosar} system and its
successor MAR~\cite{deremetza2021concept}.
This concept equips a seven-axis robotic manipulator with two HOTDOCK \cite{letier2020hotdock} interfaces
and describes a possible use case with the construction of mirror arrays in orbit.
Sharing the principal motivation behind these designs, we seek to develop a manipulator compatible with the \textit{iSSI}
interface.

A basic three-axis system was recently developed for the \textit{iSSI}~\cite{Zeis2022fully} but
is strongly limited in its capabilities to manipulate within all six Cartesian dimensions of its end-effector.
In contrast, the seven-axis of our proposed manipulator enable such
manipulation and locomotion over distributed \textit{iSSI} interfaces on the surface of
a compatible satellite.
The software needed to control such devices
is only partially covered with frameworks for space robotics applications, such as~\cite{arancon2017esrocos}.
The complexities of manipulation and locomotion shift the focus towards algorithms from
the field of robotics, such as motion planning.
It is thus of interest to evaluate open-source robotics middlewares, such as
ROS~\cite{Quigley2009}, with our use case of a locomoting manipulator.

\section{ReCoBot - Vision and Application}
\label{sec:recobot_overview}
The concept of ReCobot is embedded into the bigger picture of infrastructure
for on-orbit satellite servicing.
For such a maintenance mission, we assume that a ReCoBot-like
system is deployed on a client satellite.
We further assume that this client satellite is composed of suitable building blocks that are interconnected
by the intelligent Space System Interface (iSSI)~\cite{Kortman2015}~\cite{kreisel2019}.
To successfully perform maintenance operations,
individual blocks need to be exchanged by the robot in place.

The goal of this paper is to contribute a pragmatic, testable design for such a
system and to think through the required methods from a robotics point of view.

\section{Hardware Design}
\label{sec:hardware_design}

\begin{figure*}
    \centering
    \includegraphics[width=0.8\textwidth]{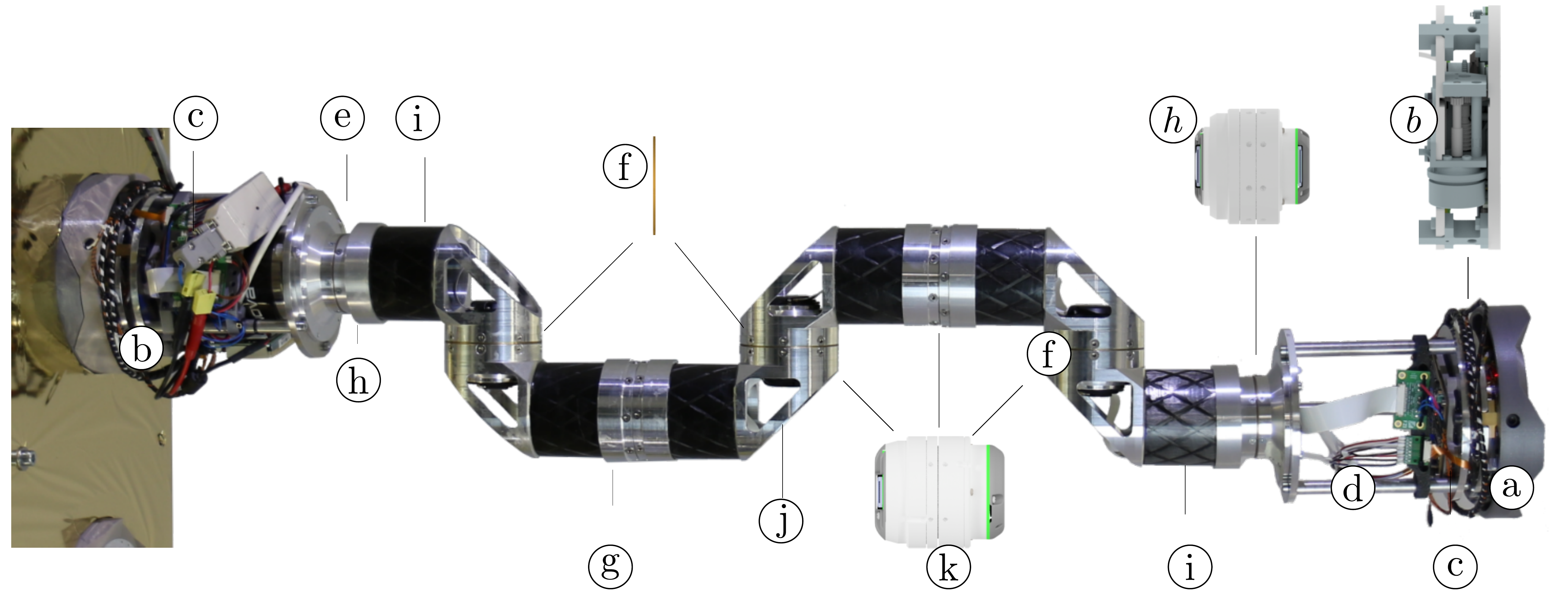}
    \caption{The fully extended ReCoBot. The letters label significant
        components of the robot. a) Form fit b) \textit{iSSI} interface
    c) \textit{ICU} adapter d) TCP housing e) Actuator adapter f) Copper
        alloy bushing g) Actuator adapter h) Actuator (KA-58) i) \textit{CFRP}
    tube j) Elbow actuator housing k) Actuator (KA-75+)
    }
    \label{fig:robot_mechanics}
\end{figure*}

ReCoBot is comprised of eight segments that are connected by seven revolute joints.
The compatibility with the \textit{iSSI} interfaces and the goal to locomote
over satellites with such block structures
determine the approximate lengths of the segments.
With this setup, the robot could reach most targets within its workspace in initial analyses.
Special emphasis was placed on a lightweight construction during the development
to ensure a functioning prototype under earth's gravity conditions.
The main materials used were aluminum alloys and carbon tubes to withstand the forces and torques under gravity.
A quasi-static estimation of reaction forces and torques in each joint was
carried out using algorithms from the Kinematics and
Dynamics Library (KDL)\footnote{http://wiki.ros.org/orocos\_kinematics\_dynamics} in ROS. 

The current ReCoBot has a maximum reach of \SI{1,20}{m} and a total mass of \SI{10,40}{kg}.

\subsection{Components}
\fig{fig:robot_mechanics} shows an overview of the complete system with details in CAD.
In the following, we list and describe the main components of the ReCoBot setup.

\begin{itemize}

\item[(a)]
Form fits at both ends of ReCoBot provide inserting chamfers for
the docking process with other satellite modules.
They shall improve the overall mechanical stability
and reduce mechanical stress of the \textit{iSSI} when docked.

\item[(b)]

The integrated \textit{iSSI} interface as described in \sect{sec:recobot_overview}.

\item[(c)](d) (g) (i) \ 
Structural components and adapters.
The TCP housing was originally designed as a closed structure, but was then
                expanded to an open, more flexible structure with three
                Aluminum alloy rods.
It accommodates the Kinova controller on one side and leaves enough space for a battery pack on the other end of ReCoBot.
Additional adapters, such as the mount for the main on-board CPU and the
                \textit{iSSI} are placed inside the TCP housing
                and are partly attached to the rods. 

\item[(e)]
Adapter for small actuators.

\item[(f)]
Copper alloy bushings between the flanges reduce friction
and absorb bending moments.
They also increase the system's overall stiffness through reducing
flex in the joints and thus mitigate uncertainty in the robot's kinematics.

\item[(h)](k) \ 
        Kinova actuators.

\item[(i)]
        \textit{CFRP} tube-shaped link segments provide structural stiffness with low weight.
        They are glued to the actuator housings.

\item[(j)]
        Elbow actuator housings are attached to the motors with screws for
        easy maintenance.

\end{itemize}

\subsection{Actuators}

We use second-generation KA-Series Kinova geared motors in ReCoBot.
Two KA-58 actuators (h) are integrated in the wrist joints, while the remaining joints are driven by five KA-75+ actuators (k). 
The KA-Series actuators consist of two disk-shaped parts, each fixed to the flange of the robotic segments, respectively.
Inside the motors, a slip ring allows for a continuous rotation\cite{rader2017highly},
which supports to lay the connecting 20 pins flat flex cable inside the segments.
The geared motors feature a torque sensor as well as an absolute position encoder.

\subsection{Optimization}
To meet the requirements of the maximum payloads of the actuators on earth, 
the elbow actuator housing (j) was subject to a series of improvements. 
\begin{figure}%
    \centering
    \subfloat[\centering Initial, \SI{687}{g} ]{{\includegraphics[width=1.6cm]{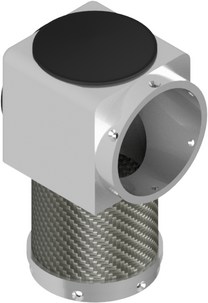} }}%
    \qquad
    \subfloat[\centering Improved, \SI{450}{g}]{{\includegraphics[width=1.6cm]{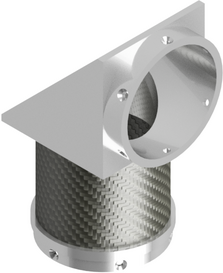} }}%
    \qquad
    \subfloat[\centering Final, \SI{315}{g}]{{\includegraphics[width=1.6cm]{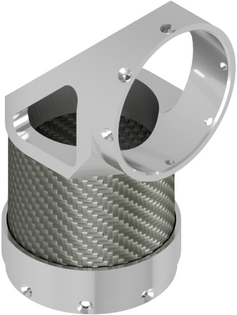} }}%
    \caption{Topology optimization of the elbow joint. The final design is a
        trade-off between light-weight and suitable to manufacture.
        }%
    \label{fig:topology}%
\end{figure}
We used a topology optimization approach to remove excess material under low
stress and evaluated the optimized design using FEM analysis.
\fig{fig:topology} shows selected steps of this refinement.
The starting point was the housing of \fig{fig:topology}(a) as a baseline. \fig{fig:topology}(b) shows an improved version, 
and \fig{fig:topology}(c) shows the final design that was created with CAD-based topology optimization.
Further improvements would be possible. Available manufacturing methods,
however, are a limiting factor in design optimization.

\section{Electronics and Integration}
\label{sec:electronics}
\begin{figure*}
    \centering
    \includegraphics[width=1\textwidth]{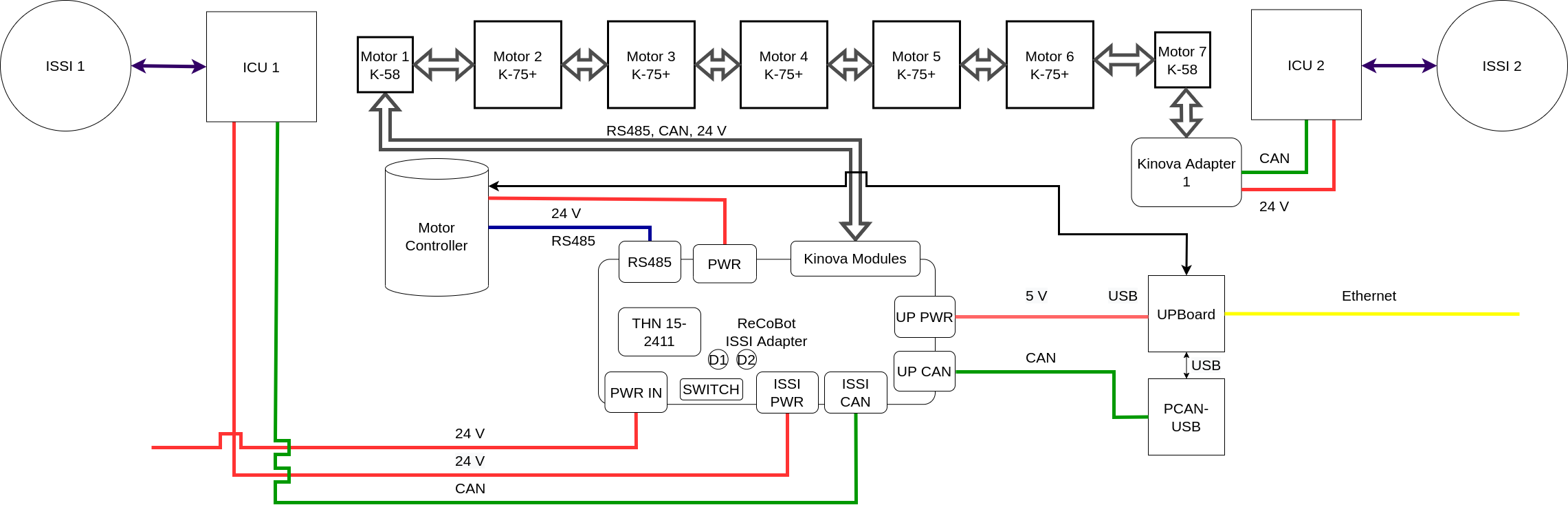}
    \caption{Electronics components and communication interfaces of ReCoBot, with
        24V power supply (red), ribbon cable (black arrow), CAN bus interface
        (green), and RS485 communication (blue).
    }
    \label{fig:electric}
\end{figure*}

One objective of the design process was to lay the cables inside the segments
along the arm to best support ReCoBot's movements.
We further designed ReCoBot to have all critical components, like the motor controller, 
the computing unit, network interfaces for Ethernet and USB, and power switches easily accessible in the open TCP housing.
Figure \ref{fig:electric} shows a schematic overview of ReCoBot's electrical components and connections.

Most of ReCoBot's electrical architecture is based on \SI{24}{V},
which is currently supplied by an external source.
Our concept envisions integrated batteries for this task at a later stage.
The \textit{iSSI} interfaces, its corresponding Interface Control Unit (ICU),
as well as the motor controller, and the KA-series actuators are operated with
this voltage. 
An additional DC/DC converter on the adapter provides \SI{5}{V} for a
compact 04/64 \textit{UP board} that serves as a single-board computer for all
of ReCoBot's software applications and control.
It runs an Ubuntu 20.04 LTS 64-bit OS with an x86 processor architecture.
The \textit{UP board} has a small mechanical footprint while providing a
performant platform for running the necessary software applications. 

The FZI ReCoBot \textit{iSSI} adapter serves as the main distribution board of the architecture.
It connects to the \textit{UP board}, the motor controller, the KA-series actuators, as well as the \textit{ICU} boards.
It provides interfaces to the power supply and the two data buses, RS485 and CAN.
The main communication line is a 20 pin flex cable, which
daisy chains the components along the robotic arm in sequential order.
\tabl{tab:table:wiring} shows the respective pin assignment.

\begin{table}
    \centering
    \caption{20 pin flat flex cable pinout}
    \label{tab:table:wiring}
    \begin{tabular}{ll}
    \hline
    Pin \# & Signal \\
    \hline
    1 - 8 & 24 V \\ 
    9 - 16 & GND \\
    17 / 18 & RS485 low / high\\
    19 / 20 & CAN low / high \\
    \hline
    \end{tabular}
\end{table}

The Kinova actuators use the RS485 specification for communication with a
dedicated input and output connector when daisy-chaining.
The FZI ReCoBot \textit{iSSI} adapter additionally splits off a CAN signal and \SI{24}{V} for controlling the \textit{iSSI} interface.
We use a \textit{PCAN} adapter for this task in combination with the \textit{UP board}.
Each \textit{iSSI} has an integrated \textit{ICU} unit.
\section{Software and Control}
\label{sec:software}
ReCoBot's software combines available open-source libraries and custom implementations
within the ROS framework~\cite{Quigley2009}.
The high-level goals of the software are
the actuation of all electronic components and control of the robot's joints;
collision-free motion planning to Cartesian target poses and joint configurations;
the locomotion over satellite structures with the \textit{iSSI} standard;
and a basic interface for teleoperation and manipulation.

A special focus is the motion planning component.
Although being established in academia and industry, \textit{Moveit}~\cite{Coleman2014reducing} requires some particular tweaks
to support our use case of locomotion, i.e. motion planning with a kinematic chain whose
basis and tip change their connection frequently with the environment.
The next sections highlight the important components and details.

\subsection{iSSI end-effectors}

The \textit{iSSI} end-effectors are controlled through a
custom CAN-bus driver, which 
encapsulates the interface with a state machine for docking ReCoBot to its environment.
In addition to opening and closing the mechanical clutches, the driver offers
continuous monitoring of internal parameters, such as supply voltage and the
temperature of critical components.
Additionally, integrated hall sensors and end stops allow for verification of
the coupled connection.

\subsection{Joint control and actuation}
Kinova's motors come with a dedicated SDK and a high-level API.
During the development for this research, however, this SDK was not compatible with
our configuration of seven Kinova motors.
We, therefore, build ReCoBot's actuation on a custom \textit{RS-485}-based driver,
which handles the communication through the Kinova motor controller.
Our driver implements a hardware abstraction layer for
and a \textit{ROS-control}\cite{Chitta2017} based hardware interface.
This enables position control at \SI{100}{Hz} with feedback about motor positions,
velocities, and a rough estimation of applied joint torques.

\subsection{Motion planning for locomotion}
\begin{figure}
        \centering
        \includegraphics[width=0.30\textwidth]{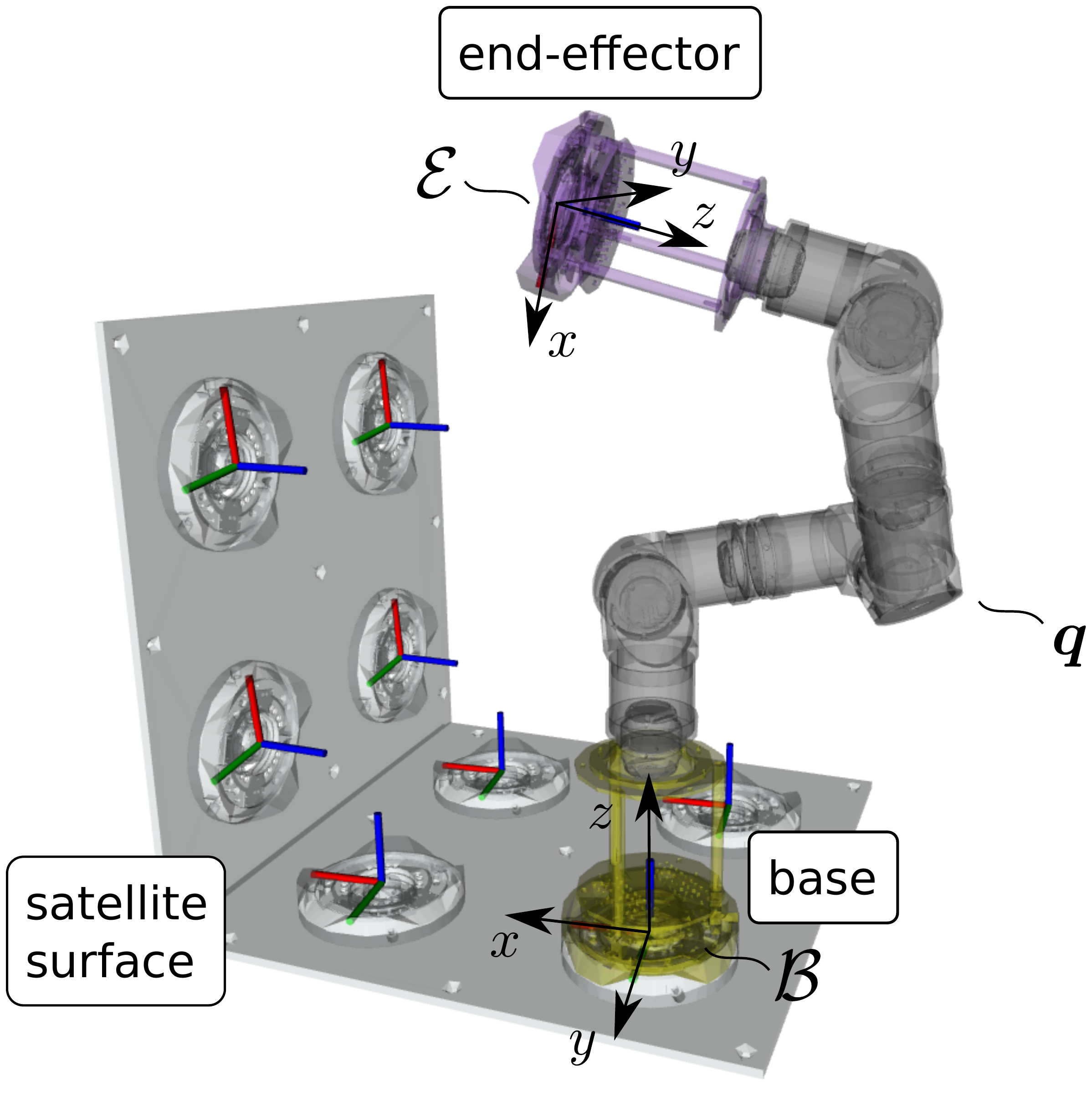}
        \caption{Exemplary planning environment for ReCoBot. The robot is
        \textit{regularly} docked via its base $\mathcal{B}$ and can approach
        satellite \textit{iSSI}s with its end-effector $\mathcal{E}$.
        }
        \label{fig:planning_environment}
\end{figure}

ReCoBot can use its \textit{iSSI}s of both ends to dock itself to satellite surfaces.
Opening and closing them dynamically allows for step-wise locomotion, and
grasping and moving individual satellite blocks can be done similarly.
We use the Moveit framework~\cite{Coleman2014reducing}
for collision-free motion planning.
Our envisioned use case of on-orbit manipulation assumes a structured environment with static collision objects.
\fig{fig:planning_environment} shows ReCoBot on the surface of an exemplary satellite in simulation.
The frames in the robot's base $\mathcal{B}$ and end-effector $\mathcal{E}$
are oriented in such a way that they coincide with the satellite's \textit{iSSI}s when docked.
All \textit{iSSI}s represent possible target poses on the satellite's surface.
We further differentiate three states for ReCoBot's configuration during locomotion:
\begin{itemize}
\item \textit{Regular docked}: The connection is closed via $\mathcal{B}$
\item \textit{Inverted docked}: The connection is closed via $\mathcal{E}$
\item In transition: Both ends are docked
\end{itemize}
Although both of ReCoBot's ends are equally capable end-effectors for
manipulation, we draw this distinction here to avoid confusion in succeeding explanations.

\begin{figure}
        \centering
        \includegraphics[width=0.49\textwidth]{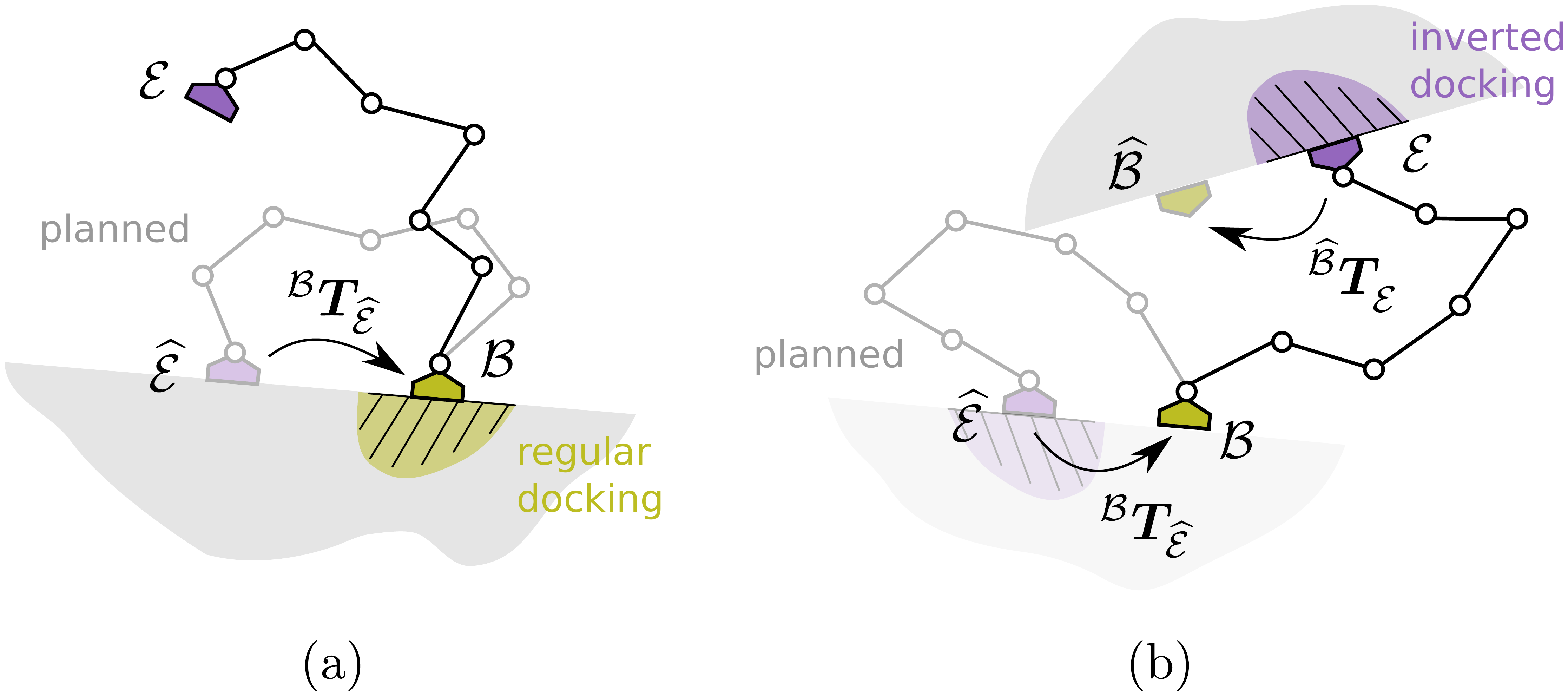}
        \caption{Schematic illustration of planning for both docking cases.
        (a) The robot is attached to the environment and reaches goal poses with its end-effector.
        (b) The environment is attached to the robot's end-effector and is
        moved inversely to reach the goal pose of the robot's base.
        }
        \label{fig:locomotion_planning}
\end{figure}

Having a calibrated model of the satellite and the robot, locomotion is
realized through a repetitive sequence of
dynamically opening and closing \textit{iSSI} connections and moving either base or tip.
This simplification allows us to use established planning algorithms.
\alg{alg:locomotion} shows the high-level steps. 
\begin{algorithm}[htbp]
        \caption{High-level steps for locomotion}
        \label{alg:locomotion}
        \begin{algorithmic}[1]
                \Procedure{Locomotion}{calibrated model}
                \State Close both of ReCoBot's \textit{iSSI}s
                \While {true}
                  \State Choose a new goal pose for $\mathcal{E}$ or $\mathcal{B}$
                  \State Open the corresponding \textit{iSSI} connection
                  \State Update the planning kinematics
                  \State Display the goal coordinates in $\mathcal{B}$.
                  \State Plan a collision-free path to this goal
                  \State Execute the motion plan
                  \State Close the \textit{iSSI} connection
                \EndWhile
        \EndProcedure
        \end{algorithmic}
\end{algorithm}
Steps $6$ and $7$ %
apply a trick to comply with the Moveit framework:
We take the robot's base $\mathcal{B}$ as the planning reference frame during locomotion
for both the \textit{regular} and \textit{inverted} docking scenarios and instead
perform a dynamic update to the robot's kinematics representation.
This allows us to use a single planning instance and smoothly transition between both docking cases.
Since we %
cannot change the given order of links and joints, we instead
need to re-attach the environment properly.
\fig{fig:locomotion_planning} illustrates both cases that we briefly describe in the following.

\subsubsection{Regular docking}
The robot kinematics is attached to the environment via its base $\mathcal{B}$ and
the planned goal pose of the end-effector is given with $\widehat{\mathcal{E}}$.
Its coordinates are determined by the transformation ${}^{\mathcal{B}}
\bm{T}_{\widehat{\mathcal{E}}}$ that transforms from goal coordinates to the
planning reference.
\fig{fig:locomotion_planning}(a) shows an exemplary planning result.
After trajectory execution, the robot's end-effector $\mathcal{E}$ coincides
with the planned goal pose $\widehat{\mathcal{E}}$.

\subsubsection{Inverted docking}
The environment is attached to the robot's kinematics via the end-effector $\mathcal{E}$.
\fig{fig:locomotion_planning}(b) illustrates an exemplary planning goal $\widehat{\mathcal{B}}$ for the robot's base.
We obtain the coordinates from the transformation ${}^{\widehat{\mathcal{B}}} \bm{T}_{\mathcal{E}}$,
which is equivalent to describing the end-effector goal
$\widehat{\mathcal{E}}$ after the planned execution with respect to the robot's
planning reference $\mathcal{B}$.
After trajectory execution, the robot's end-effector has moved the
environment in such a way that the robot's base $\mathcal{B}$ coincides with
the planned goal $\widehat{\mathcal{B}}$.

We obtain all transformations as lookups from our calibrated environment that
incorporates real-time feedback via forward kinematics from ReCoBot's joint states.
During locomotion, each planning result is a
smooth trajectory with a user-defined duration that is executed
with the joint trajectory controller.

\subsection{Cartesian control and teleoperation}
\begin{figure}
        \centering
        \includegraphics[width=0.49\textwidth]{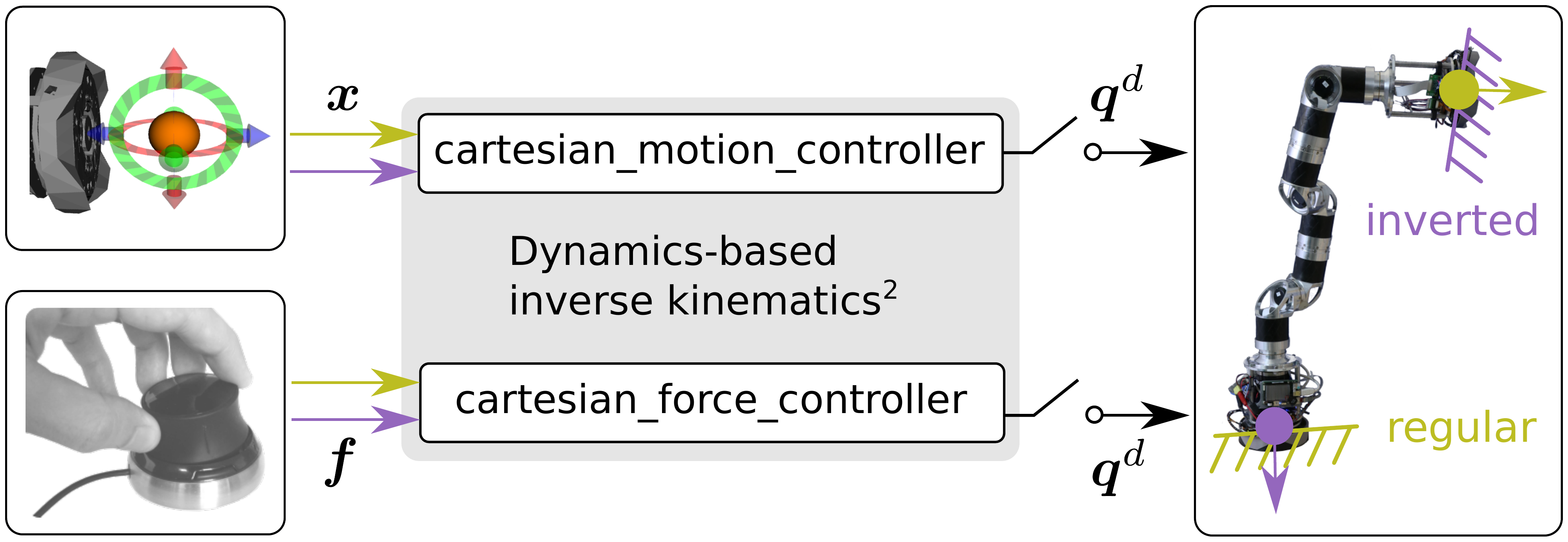}
        \caption{Cartesian control and teleoperation.
        Two manual control interfaces allow steering ReCoBot's end-effector in
        both the \textit{regular} and the \textit{inverted} docking scenario.
        }
        \label{fig:cartesian_control_and_teleoperation}
\end{figure}

Collision-free motion planning covers
broad workspace motions with arbitrary configuration changes.
The goal of the additional, manual control interfaces is to offer an
intuitive alternative for fine-grained manipulation in a relatively small volume.
Offering a suitable trade-off between simplicity and intuitiveness,
it is assumed that no configuration changes and no self-collision
checking are necessary in this case.
\fig{fig:cartesian_control_and_teleoperation} shows an overview.

The two controller types \textit{cartesian\_motion\_controller} and
\textit{cartesian\_force\_controller} are used from our control
suite\footnote{https://github.com/fzi-forschungszentrum-informatik/cartesian\_controllers}.
They turn task space inputs into position commands for the seven joints.
Two separate instances of each controller can be used to steer the
tip of the kinematic chain in the \textit{regular} and \textit{inverted}
docking case, respectively.

Both controllers share a common solver for the inverse kinematics problem that
computes joint position commands with a dynamics-based approach.
In short, this solver turns user input into a virtual, goal-directed vector that
drives a virtual model of the robot in that direction.
The simulated reaction of the system is then streamed as desired
reference $\bm{q}^d \in \mathbb{R}^7$ to the joint position controllers of the real robot.
A feature of the solver is a good trade-off between task space linearity and singularity robustness through
conditioning the virtual operational space inertia of the system.
We refer the interested reader to \cite{Scherzinger2020virtual} and \cite{Scherzinger2019Inverse} for more details on this concept.

The Cartesian control supports two input modalities to steer ReCoBot:
The first one is a virtual 6D handle in our visualization environment
that formulates the end-effector's target pose $\bm{x} \in \mathbb{R}^7$
(position and rotation quaternion) with respect to the base frame of the
current docking.
Its strength is precise control in individual axes.
The second input modality is
more intuitive for steering several Cartesian axes at the same time.
We use a \textit{3DConnexion} spacemouse as a joystick which offers to measure displacements in all six Cartesian axes.
The data are adequately scaled according to the desired manipulation sensitivity and then
passed to the \textit{cartesian\_force\_controller} in form of a force-torque vector $\bm{f} \in \mathbb{R}^6$.
The term force control, however, needs a brief explanation:
Reaction forces with the environment are currently not actively controlled and
$\bm{f}$ is purely used for steering the robot.
This is due to the quality of measured joint torques being currently insufficient for Cartesian closed-loop control.
Interaction with the environment must thus be anticipated by the operator,
which can be realized through observing the rendering of real-time joint feedback for instance.
A minimal amount of compliance for slow motion in contact is provided by the structural mechanics.

\section{Evaluation and Results}
\label{sec:evaluation}

We tested ReCoBot's core functionalities both in a basic simulator and on a specially designed mockup in a lab environment.
The simulator has the advantage of providing an idealized behavior for an in-depth
analysis of the used algorithms, and was used for evaluating the locomotion concept individually.
In addition, the lab environment allows for evaluating ReCoBot's hardware and electronics concept within focussed use cases.
The next sections provide an excerpt of two examples with highlights.

\subsection{Software and Locomotion}
\begin{figure}
        \centering
        \includegraphics[width=0.47\textwidth]{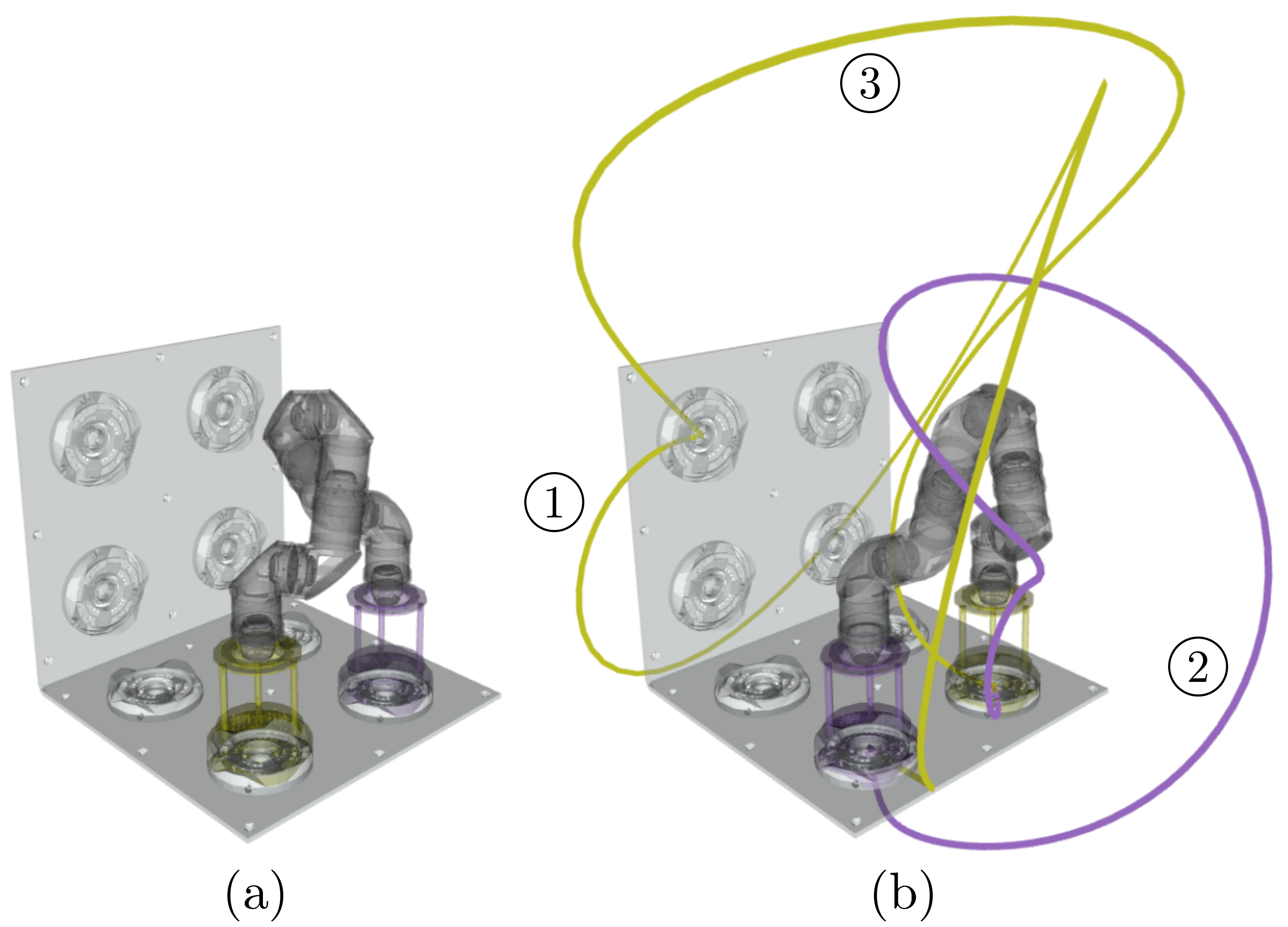}
        \caption{Visualization of planning and locomotion. ReCoBot switches its base and
        end-effector with the help of an \textit{iSSI} interface on the top-left corner.
        }
        \label{fig:evaluation_planning}
\end{figure}

The basic simulation environment consisted of a kinematics representation of ReCoBot, docked to the surface dummy of
an exemplary satellite.
Dynamic phenomena were neglected and the robot's joint control simply reported back the commanded positions as current states.
The motion planning component had a sufficiently exact knowledge of the robot
and its environment in form of approximated collision meshes from CAD data.
The robot started in a transition phase, as depicted in \fig{fig:evaluation_planning}(a),
and the goal was to switch base and end-effector in place.

\fig{fig:evaluation_planning}(b) shows the resulting locomotion in form of traveled end-effector paths.
The endpoint traces were computed based on forward joint position kinematics
and the colors correspond to the \textit{regular} and \textit{inverted} docking case from \sect{sec:software}.
In this example, the robot had to utilize a user-specified interface on the top left corner to relocate its base.
The numbers next to the recorded paths show the order of execution.
We used the \textit{RRT${}^*$} algorithm~\cite{Karaman2011sampling} for these experiments,
which found quick initial solutions and used the given planning time of
\SI{20}{s} per movement for further optimization of path costs.

We found through the experiments that most target poses where reachable within the given grid of \textit{iSSI}s.
We noted, however, that the high mesh accuracy of the form fits negatively
influenced the success rate of the planners due to recognized collisions in the
robot's start and end states, which could be circumvented with a specific adaptation
of collision meshes for motion planning.

\subsection{System Tests in the Lab Environment}

\begin{figure*}
        \centering
        \begin{subfigure}[b]{0.16\textwidth}
                \includegraphics[width=1.0\textwidth]{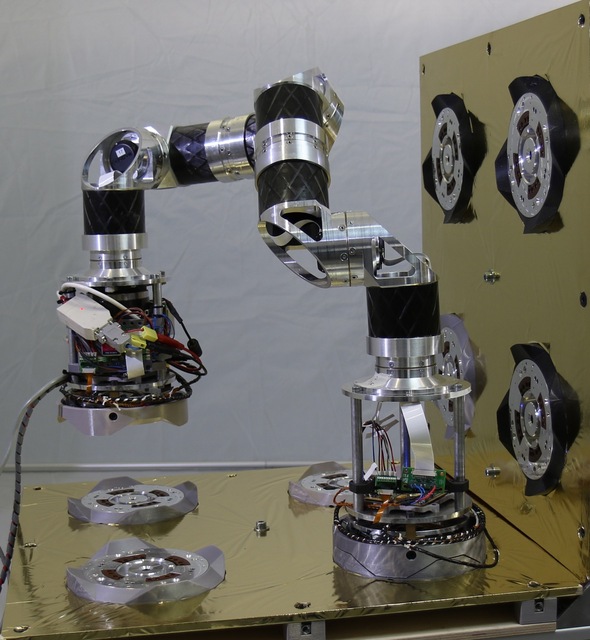}
                \caption{}
        \end{subfigure}~%
        \begin{subfigure}[b]{0.16\textwidth}
                \includegraphics[width=1.0\textwidth]{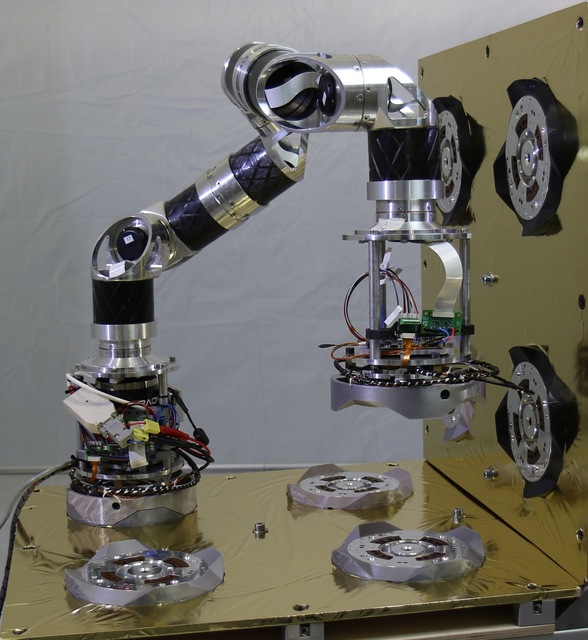}
                \caption{}
        \end{subfigure}~%
        \begin{subfigure}[b]{0.16\textwidth}
                \includegraphics[width=1.0\textwidth]{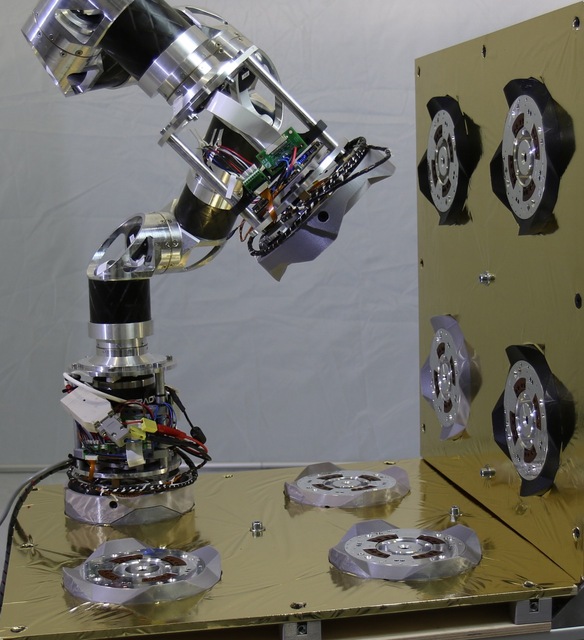}
                \caption{}
        \end{subfigure}~%
        \begin{subfigure}[b]{0.16\textwidth}
                \includegraphics[width=1.0\textwidth]{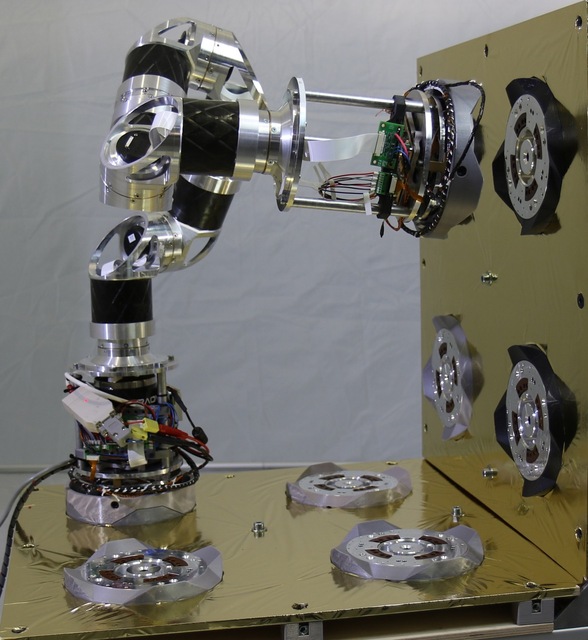}
                \caption{}
        \end{subfigure}~%
        \begin{subfigure}[b]{0.16\textwidth}
                \includegraphics[width=1.0\textwidth]{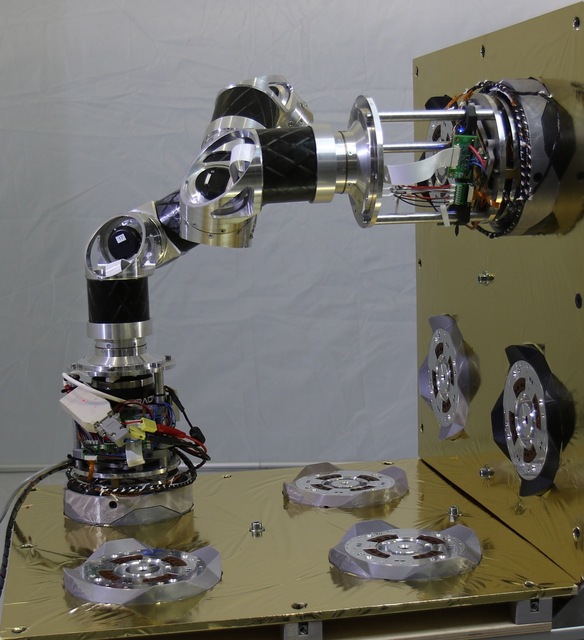}
                \caption{}
        \end{subfigure}~%
        \begin{subfigure}[b]{0.16\textwidth}
                \includegraphics[width=1.0\textwidth]{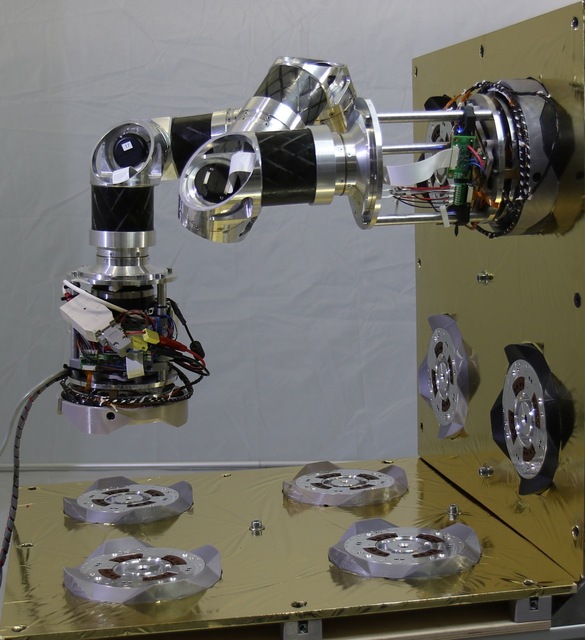}
                \caption{}
        \end{subfigure}~%
        \caption{Tests with ReCoBot on a satellite mockup. (a)-(b) repetitive
        opening and closing of interfaces. (c)-(e) vertical docking at a
        distant interface. (f) Teleoperation of the robot's base along the
        vertical axes.
        }
        \label{fig:lab_tests}
\end{figure*}

The mockup consisted of two rectangular-oriented walls with
eight \textit{iSSI}s arranged in a grid.
Similar to the simulation environment, the \textit{iSSI} centers were arranged \SI{30}{cm} apart.

We tested ReCoBot and the interplay of all components in different docking scenarios.
The joints and \textit{iSSI} end-effectors were actuated by the integrated onboard PC and the robot published
real-time feedback over LAN for further processing.
The \textit{iSSI} interfaces of the dummy satellite were not actuated and provided the
passive counterpart for the active bayonet clutch of the robot's ends.
The opening and closing of the end-effector \textit{iSSI}s for docking were triggered by user commands.
For these experiments, we powered the robot through an external cable.

\fig{fig:lab_tests}(a)-(b) show an alternation
between a \textit{regular} and an \textit{inverted} configuration and a
repetitive opening and closing of the \textit{iSSI} interfaces.
The tests could be executed autonomously and reproducibly on the horizontal
parts of the mockup after a calibration of the robot to the environment.
We found that the form fits could compensate only very small offsets during joint position control.
When working with uncalibrated environments, active compliance by control will be necessary to compensate
for the uncertainty.

\fig{fig:lab_tests}(c)-(e) show the approach and the vertical docking of a more distant interface.
We needed a manual calibration of the final docking poses in these cases due to
a joint-dependent, non-linear structural compliance of the robot under gravity, which leads to
individual offsets for each \textit{iSSI}.
The robot's joint arrangement allowed it to reach these interfaces well.

\fig{fig:lab_tests}(f) shows ReCoBot lifting its base to transition into an \textit{inverted} docking.
We teleoperated the robot via the virtual 6D handle and could steer the
end-effector with sufficient accuracy in Cartesian axes.

In our experiments, we frequently reached the supported torque limits of the
smaller motors in vertical docking scenarios, even with comparatively slow execution speeds.
This torque limitation together with the gravity-induced uncertainty,
however, will play no role in the envisioned zero-gravity environment of on-orbit
manipulation.
\section{Conclusions and Future Works}
\label{sec:conclusions}

This paper presented the ReCoBot, a seven-axis robotic manipulator for on-orbit locomotion
on \textit{iSSI} compatible satellites.
The system uses two end-effector mounted \textit{iSSI} interfaces for docking
and realizes locomotion planning and teleoperation with Moveit
and ROS-control.
Experiments evaluated the validity of the proposed concepts for mechanics and electronics in docking maneuvers on a mockup.
Future work could consider torque optimal planning~\cite{Kalakrishnan2011stomp} for reduced energy consumption.
The integration of additional end-effector force-torque sensors and active compliant control could
compensate for uncertainty with force-sensitive docking.
We will further investigate the possibility of space qualification of ReCoBot in its current state,
and consider software-side safety and redundancy through a switch to
ROS2~\cite{Maruyama2016} with real-time context~\cite{Puck2020distributed}.
\renewcommand*{\bibfont}{\small}

\printbibliography

\end{document}